\documentclass[letterpaper, 10 pt, conference]{ieeeconf}

\usepackage{times}
\usepackage{epsfig}
\usepackage{graphicx}
\usepackage{amsmath}
\usepackage{amssymb}
\usepackage{graphics} 
\usepackage{epsfig} 
\usepackage{mathtools}
\usepackage{subcaption}
\usepackage{multirow}
\usepackage[table]{xcolor}
\usepackage{tabularx}
\makeatletter
\let\NAT@parse\undefined
\makeatother
\usepackage{hyperref}
\hypersetup{
    colorlinks=true,
    linkcolor=blue,
    filecolor=magenta,      
    urlcolor=cyan,
    pdftitle={RAIST},
    pdfpagemode=FullScreen,
    }

\usepackage[font=small,belowskip=-1.5pt]{caption} 

\newcommand{\namedref}[2]{\hyperref[#2]{#1~\ref*{#2}}}

\newcommand{\p}{\mathbf{p}}

\newcommand{\app}{\mathbf{a}}
\newcommand{\ro}{\mathbf{r}}
\newcommand{\context}{\mathbf{c}}
\newcommand{\G}{\mathbf{G}}

\newcommand{\X}{\mathbf{X}}
\newcommand{\W}{\mathbf{W}}
\newcommand{\T}{\Gamma}
\newcommand{\I}{\mathbb{I}}
\newcommand{\R}{\mathbb{R}}
\newcommand{\Img}{\mathcal{I}}
\newcommand{\argmax}{\mathop{\mathrm{argmax}}}

\DeclarePairedDelimiter\floor{\lfloor}{\rfloor}

\title{\LARGE \bf \textit{RAIST}: Learning Risk Aware Traffic Interactions via Spatio-Temporal Graph Convolutional Networks}
\author{Videsh Suman, Phu Pham and Aniket Bera \\
{Department of Computer Science, Purdue University, USA}\\
\vspace{-15pt}
}

\begin{document}

\maketitle
\thispagestyle{empty}
\pagestyle{empty}

\begin{abstract}
A key aspect of driving a road vehicle is to interact with other road users, assess their intentions and make risk-aware tactical decisions. An intuitive approach to enabling an intelligent automated driving system would be incorporating some aspects of human driving behavior. To this end, we propose a novel driving framework for egocentric views based on spatio-temporal traffic graphs. The traffic graphs model not only the spatial interactions amongst the road users but also their individual intentions through temporally associated message passing. We leverage a spatio-temporal graph convolutional network (ST-GCN) to train the graph edges. These edges are formulated using parameterized functions of 3D positions and scene-aware appearance features of road agents. Along with tactical behavior prediction, it is crucial to evaluate the risk-assessing ability of the proposed framework. We claim that our framework learns risk-aware representations by improving on the task of risk object identification, especially in identifying objects with vulnerable interactions like pedestrians and cyclists.
\end{abstract}
\section{Introduction}



Over the years, consistent research efforts toward autonomous driving have led to significant developments in perception, planning, and control. One major challenge for autonomous vehicles is ensuring safe planning and driving decisions in urban traffic. To achieve this, robust methods are required for accurately and efficiently understanding traffic, similar to human drivers, and for analyzing complex traffic scenes to make tactical driving decisions such as slowing down, changing lanes, or taking turns. Making these decisions recklessly in moving traffic can lead to road accidents and fatalities, which makes it critical to learn how to identify potential risks in driving scenarios.

While visual cues are often the most useful source of perception and holistic scene understanding, utilizing this information to assess the behavior and interaction of individual road users is not straightforward. For urban driving scene understanding, it is necessary to anticipate the behavior of dynamic and heterogeneous road agents in traffic, including vulnerable road users like pedestrians and cyclists, as well as other motorized road agents. Recent works \cite{li2020learning, zhang2020interaction} have shown promise in modeling spatial interactions among road users using interaction graphs. However, the current methods fail to adequately model the temporal behavior of individual road agents. Anticipating the behavior of specific road agents is crucial for safe human driving and is equally critical for autonomous vehicles to plan navigation in impending traffic scenarios.

Temporal intention modeling, for instance, can help the vehicle distinguish between a scenario with a pedestrian waiting to cross the road and another with a pedestrian walking parallel to it. Another example where temporal modeling is beneficial is in anticipating the intent of an approaching vehicle at a road intersection, i.e., predicting the direction in which the vehicle will continue moving.

Apart from aiding driving decisions, a deeper understanding of the scene, including the intention modeling of road agents, can also be leveraged to assess the risk involved. However, deep learning models are often uninterpretable and can fail unexpectedly. An auxiliary goal of risk assessment is to improve interpretability in predicting driver-centric tactical behavior. The risk assessment literature \cite{lefevre2014survey} extensively covers potential risk identification, but current methods, such as estimating object importance \cite{gao2019goal, zhang2020interaction}, require carefully annotated training labels. Another plausible approach is to identify \textit{risk objects} that causally influence ego-behavior \cite{li2020make}, but existing methods lack sufficient interaction modeling and perform poorly, particularly concerning pedestrians.




\begin{figure}[h]
  \centering
  {\includegraphics[width=1\linewidth]{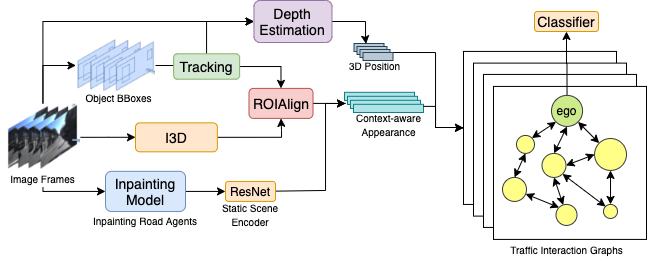}}%
  \caption{\textit{An overview of the proposed driving model. The model takes a sequence of video frames as input, applies 3D convolutions (I3D backbone) to extract visual features, and employs ROIAlign to extract appearance features. Depth estimation and inpainting modules are used to obtain object-level 3D positions and inpainted frames with the background scenes, respectively. A static scene encoder is used to encode the inpainted images to obtain the context, which is further combined with the object-level appearance features. The modified features, along with the object-level 3D positions, are used to construct a frame-wise traffic interaction graph learned via spatio-temporal graph convolutional networks.}}
  \label{fig:framework}
\end{figure}

\textbf{Main Contributions:} In this paper, we present a novel spatio-temporal learning framework for causal risk assessment using egocentric videos. We demonstrate that our proposed framework can generalize to multiple tasks that benefit from interactive driving scene understanding. Our approach involves utilizing a combination of detection, tracking, depth estimation, and action recognition to obtain the attributes of road agents, which are then used to formulate the edges of spatio-temporal graphs. These graphs are instrumental in computing a risk metric. The main contributions of this work are as follows:
\begin{itemize}
    \item We employ object attributes to model interaction edges among different object nodes at a particular instant. Each object node is also associated with its own representations across multiple time steps through temporal message passing. The interaction edges and temporal message passing are parameterized with respect to the vulnerability of the participating road agents.
    \item We employ inpainting to remove dynamic road agents from the input image frames, followed by feature extraction, to generate a frame-wise \textit{static scene prior}. This scene prior is combined with each object's appearance features for context awareness, which is ultimately used to model the graph edges.
    \item We evaluate the proposed approach on the task of risk object identification in a label-efficient manner using causal inference \cite{li2020make}. The advantages of using this framework to solve the causal problem are detailed in Section \ref{sec:risk_assessment}.
    
\end{itemize}
\vspace{-10pt}


\section{Related Work}

\subsection{Driving Scene Understanding}
There have been significant research efforts in driver behavior understanding and tactical behavior recognition \cite{oliver2000graphical},  \cite{wu2013reasoning}. Classical approaches often employed hidden Markov models (HMM) to recognize driver behaviors, where a single node in the HMM represented states from the ego vehicle, roads, and traffic participants in a state vector. However, in our proposed framework, we explicitly focus on modeling interactions between the ego vehicle and other traffic agents, encompassing two of these three states. The third state, representing roads, is implicitly captured using object-level context-aware features obtained from the static scene encoding. 

In the era of deep learning, several interesting methods \cite{li2020learning}, \cite{mylavarapu2020understanding},  \cite{RABEV} have been proposed to address similar challenges. While traffic trajectory prediction \cite{chandra2019traphic, chandra2019robusttp, chandra2020forecasting} and tracking \cite{chandra2019densepeds, chandra2019roadtrack, chandra2019graphrqi, cmetric} have been widely studied in robotics and computer vision, driver behavior modeling has mostly been confined to traffic psychology and the social sciences \cite{chandra2020stylepredict,ishibashi2007indices,ernestref2,ernestref3,ernestref4,ernestref8,ernestref9}.

\subsection{Graph Convolutional Networks for Driving Scenes}
Among various deep learning methods, graph convolutional networks \cite{kipf2016semi} have demonstrated significant progress in tasks such as action recognition \cite{yan2018spatial}, group activity recognition \cite{wu2019learning, timothyha}. These networks excel in learning the underlying semantic structure in input videos, making them well-suited for autonomous driving applications.

Recent approaches \cite{li2020learning}, \cite{mylavarapu2020understanding}, \cite{zhang2020interaction} have proposed using graph convolutions to learn the spatial semantics of ego-centric driving scenarios. However, none of these works have focused on object-level temporal intention modeling using message passing via temporally associated nodes. To address this gap, we aim to leverage spatio-temporal GCNs \cite{yan2018spatial} in our framework.

\subsection{Risk Assessment}
Risk assessment is pivotal in intelligent vehicle decision-making. Current methodologies, as reviewed in \cite{lefevre2014survey}, define risk as the misalignment between expectations and actual traffic scenarios, a definition echoed in \cite{li2020make}. In our model, we aim to assess risk based on this definition. While closely tied to the task of estimating object importance in driving scenes \cite{gao2019goal}, \cite{zhang2020interaction}, no previous methods have specifically evaluated the vulnerability of traffic participants to the ego vehicle, which our framework intends to address.
\section{Methodology}

This section describes our risk-aware driving scene understanding algorithm, the feature extraction pipeline, and the spatio-temporal graph-based approach.

\subsection{Problem Statement}
We present a comprehensive framework designed to predict tactical driver behavior and perform causal reasoning from driver-centric traffic, enabling causal risk assessment using egocentric videos. In this context, we define \textit{tactical driver behavior} as the high-level driving decision resulting from a \textit{cause} or the action of one or more participating dynamic road agents. For instance, a tactical behavior could involve \textit{changing lanes}, with the corresponding cause being a \textit{parked vehicle}. To address this challenge, we formulate it as a classification problem with a predefined set of tactical behaviors and their corresponding causes. Our framework capitalizes on the underlying structure of a traffic scenario, which includes the heterogeneity in traffic agents. To achieve this, we extract semantic features of traffic agents and utilize them to learn pair-wise spatial interactions that evolve over time. This enables us to build a more generalizable and interpretable model for driving scene understanding.

\subsection{Feature Extraction}
In this section, we elaborate on the multiple feature extraction steps based on the semantics of the static and dynamic traffic scenarios.

\textbf{Tracking and Location:} Starting with a sequence of $\T$ consecutive RGB frames denoted as $\Img^{1:\T}$, we employ Faster R-CNN~\cite{ren2015faster} to detect road agents in each frame. To uniquely identify each road agent across time, we use Deep SORT~\cite{wojke2017simple} to associate the obtained detections, resulting in $N$ unique road agents represented as \textit{tracklets}. Each tracklet, denoted as $\ro_i$, indicates the presence or absence of a road agent throughout the $\T$ time steps. Due to perspective projection, directly using the 2D pixel positions of objects for modeling spatial relations among them can be challenging. To address this, we perform depth estimation \cite{lasinger2019towards}, followed by inverse projection, which allows us to transform each object's 2D pixel position into a 3D space relative to the camera frame.
\begin{align}
\begin{bmatrix} 
x & y & z
\end{bmatrix}^T
= \delta_{u, v}  \mathbf{K}^{-1} 
\begin{bmatrix}
u & v & 1
\end{bmatrix}^T
\end{align}
where $(u, v)$ and $(x, y, z)$ represents a point in 2D and 3D coordinate systems respectively.
$\mathbf{K}$ is the camera intrinsic matrix and $\delta_{u, v}$ is the relative depth at $(u, v)$ obtained by depth estimation \cite{lasinger2019towards}. In the image plane, we select the center of a bounding box as that object's 2D position.  

\textbf{Appearance:} To perform spatio-temporal reasoning on the road agents, we need rich features that are able to encode their inherent characteristics across all $\T$ time steps from the input $\Img^{1:\T}$. For this, we use Inception V1~\cite{szegedy2015going} and I3D~\cite{Carreira_2017_CVPR} as the global video descriptors and apply ROIAlign~\cite{he2017mask} over the 3D video feature map to obtain object-level appearance features from the corresponding regions of interest.

\textbf{Static Scene Encoding:} The appearance of road agents should also be associated with their static environment. To this end, we formulate a context extraction module in our framework. With $\Img^{1:\T}$  as the input, this module performs inpainting ~\cite{yu2018generative} to generate $\mathcal{S}^{1:\T}$ image frames free from the dynamic agents of the scene. Following the inpainting, we use ResNet-18 \cite{he2016deep} to encode $\mathcal{S}^{1:\T}$ into frame-wise context embeddings $\context^{1:\T}$. We update each object's appearance feature $\app^t$ for time step $t$, 
\begin{align}
    \app^t \leftarrow \app^t \odot \context^t
\end{align}

\textbf{Vulnerability:} Vulnerable road agents are those that have little or no external protection and are at most risk in traffic~\cite{constant2010protecting}. In our design, we consider detected objects of classes \textit{person, bicycle, car, motorcycle, bus, truck} as road agents, categorizing \textit{person, bicycle} as \textbf{\textit{vulnerable}} $(m = 1)$ and the rest as non-vulnerable $(m = 0)$. We use this categorization to learn a separate set of parameters for vulnerable road agents in relation modeling.

In summary, we obtain $N$ road agents and their frame-wise attributes from the input $\Img^{1:\T}$. For each road agent, we have (a) its tracklet $\ro_i$, and if it exists in frame $\Img^{t}$, we have its (b) category label $m^t_i$, (c) spatial position $\p^t_i$ in 3D, and (d) scene-aware appearance feature $\app^t_i$. To perform spatio-temporal learning on graphs, we use the three mentioned attributes of road users.

\subsection{Spatio-Temporal Modeling}
The extracted features of road agents are used to create spatial interaction graphs with parameterized edges. We use spatio-temporal graph convolutions to learn the desired representations towards traffic scene understanding. We discuss this in detail here.

\textbf{Graph Definition:}  
In correspondence with $\Img_{1:\T}$, let $\G_{1:\T}$ be the sequence of adjacency matrices, each of which can be realized as a graph with weighted edges between all existing nodes at that time step. Such a spatial graph $\G_t$ can be interpreted as a traffic scene $\Img_t$, where element $\G_t(i, j)$ denotes the influence of agent $v^t_i$ over agent $v^t_j$. In any spatial graph $\G_t$, each node $v^t_i$ can be represented as a tuple of its attributes $\{(\app_i^t, \p_i^t, m_i^t) | i = 1, 2, \dots, K+1\}$, assuming there are $K$ road agents excluding the \textit{ego}, a non-vulnerable road agent that is spatially located at the origin in the camera frame and whose context-aware appearance feature is based on the global output feature from the video encoder I3D.

\textbf{Spatial Interaction Modeling:}
We believe that the interaction between any two road agents should be unique and depend on their respective attributes. To this end, we formulate the pair-wise spatial interactions using appearance and positional relations based on~\cite{wu2019learning}. 

We represent the interaction value between nodes $v^t_i$ and $v^t_j$ with an entry in the adjacency matrix:
\begin{align}
    \G_t(i, j) = \frac{\I_d(\p_i^t, \p_j^t) f_p(\p_i^t, m_i^t, \p_j^t, m_j^t) \exp{(f_a(\app_i^t, \app_j^t))}}{\sum_{j=1}^{K+1} \I_d(\p_i^t, \p_j^t) f_p(\p_i^t, m_i^t, \p_j^t, m_j^t) \exp{(f_a(\app_i^t, \app_j^t))}}
\end{align}
where $\I_d$ indicates the distance constraint, $f_p(\p_i^t, m_i^t, \p_j^t, m_j^t)$ indicates the vulnerability-dependent positional relation and $f_a(\app_i^t, \app_j^t)$ indicates the context-aware appearance relation. For every node, its edges are normalized, resembling a weighted softmax function. The influence of $v^t_i$ over $v^t_j$ is represented as a fraction of the total influence of $v^t_i$. The adjacency matrix $\G_t$ also includes self-connections for self-attentional learning.

In order to restrict the interaction edges to objects that are spatially closer in traffic, we use the distance constraint as:
\begin{align}
    \I_d(\p_i^t, \p_j^t) = \I(d(\p_i^t, \p_i^t) \leq \mu)
\end{align}
where $\I(.)$ is the indicator function, $d(\p_i^t, \p_j^t)$ is the Euclidean 3D distance between objects $v^t_i$ and $v^t_j$, and $\mu$ is the distance threshold that constrains the interaction $\G_t(i, j)$ to be non-zero only if the spatial distance is lower than this upper bound. We use $\mu=3$ in our framework.

The appearance relation is based on mutual similarity~\cite{vaswani2017attention} and calculated as the scaled dot product between two embeddings as:
\begin{align}
\vspace{-30pt}
    f_a(\app_i^t, \app_j^t) = \frac{\phi(\app_i^t)^T \omega(\app_j^t)}{\sqrt{D}}
\vspace{-20pt}
\end{align}
where $\phi(.)$ and $\omega(.)$ are learnable linear transformations over the appearance features and $D$ is the dimension of the embedding space. The parameters are learned to compute a scalar value for the influence of $v^t_i$ over $v^t_j$ in the embedding space. We note that this relation makes the overall spatial interaction directional, as $\phi(.)$ and $\omega(.)$ are different transformations. 

For positional relation encoding, we use a vulnerability-dependent learnable transformation function $\theta_{m_i^t}(\p_i^t)$ to embed 3D positions. The pair-wise positional relation can be formulated as:
\begin{align}
    f_p(\p_i^t, m_i^t, \p_j^t, m_j^t) = \text{ReLU}\left(\W_p\gamma\left(\theta_{m_i^t}(\p_i^t) \odot \theta_{m_j^t}(\p_j^t)\right)\right)
\vspace{-20pt}
\end{align}
where $\odot$ denotes element-wise sum, and $\gamma(.)$ is the Gaussian mapping function which encodes the embedded input into a high dimensional representation using sine and cosine of different wavelengths. 
\begin{align}
    \gamma(\mathbf{v}) = [cos(2\pi\mathbf{Bv}), sin(2\pi\mathbf{Bv})]^T
\end{align}
where $\mathbf{v} \in \mathbb{R}^d$ is the input embedding vector and $\mathbf{B} \in \mathbb{R}^{k\times d}$ consists of entries sampled from $\mathcal{N}(0, \sigma^2)$, and $\sigma$ is a hyperparameter. Here, $\gamma(.)$ is used to embed $d$-dimensional input vector $\mathbf{v}$ into $2k$-dimensional vector.
Such a Fourier feature mapping is effective in overcoming the spectral bias inherent in learning high-frequency functions in low dimensional domains \cite{tancik2020fourier} like the positional relation in our case. This mapping is then transformed into a scalar using trainable $\mathbf{W}_p$, followed by ReLU activation for non-linearity. 

\textbf{Spatio-Temporal Graph Reasoning:}
With the interaction graphs $\G_{1:\T}$ formulated, we use spatio-temporal graph convolutional network (ST-GCN) \cite{yan2018spatial} as the graph learning module over them. ST-GCN follows a similar implementation as GCN \cite{kipf2016semi}, which is followed by a convolution in the temporal domain. A layer of ST-GCN takes features as the input, leverages the spatial edges to aggregate the node-wise influence, and then uses spatial and temporal convolutional kernels to update the input features. For the purpose of simplification, we express $l$-th layer of ST-GCN at a particular time-step $t$,
\begin{align}
    \X^{l'}_t &= \text{ReLU}(\G_t\X^l_t\W^a_t) \label{eqn:scn} \\
    \X^{l+1}_t &= \text{ReLU}(\sum_{i = -\floor{\tau/2}}^{{\floor{\tau/2}}}\X^{l'}_{t+i}\W^b_{t+i} \ + \ \X^l_t) \label{eqn:tcn}
\end{align}
where $\G_t \in \R^{N\times N}$ is the graph adjacency matrix, $\X^l_t \in \R^{N\times C}$ is $l$-th layer's feature map, ReLU is the activation function, $\W^a_t \in \R^{C\times C}$ and $\W^b_t \in \R^{C\times C}$ are the weight matrices for transformations in spatial and temporal dimensions respectively. In Equation \eqref{eqn:tcn}, spatially attended interaction features are aggregated over $\tau$ consecutive time steps. There is a residual connection in each layer. In practice, this pair of transformations is implemented via 2D convolutional layers on a stack of frames in parallel, and the temporal span $\tau$ is used as the size of the temporally convolving kernel.

In the case of traffic scenarios, road agents keep varying across time. For effective temporal message passing, we need their associativity across $\T$. In practice, the spatio-temporal adjacency matrix $\G_{1:\T} \in \R^{\T\times N \times N}$, and all the invalid interactions are zeroed based on the information from the tracklets $\ro_{1:N}$. We use the context-aware appearance feature map stacked temporally as the input to the ST-GCN layers and update them after every layer. Following this, the interaction edges are remodeled with the updated features to be used for the next layer of ST-GCN. For learning the bias towards vulnerable road agents, we assign separate temporal convolutional weights to be shared across vulnerable and non-vulnerable agents.


\section{Risk Assessment} \label{sec:risk_assessment}

\begin{figure}[!t]
  \centering
  {\includegraphics[width=1\linewidth]{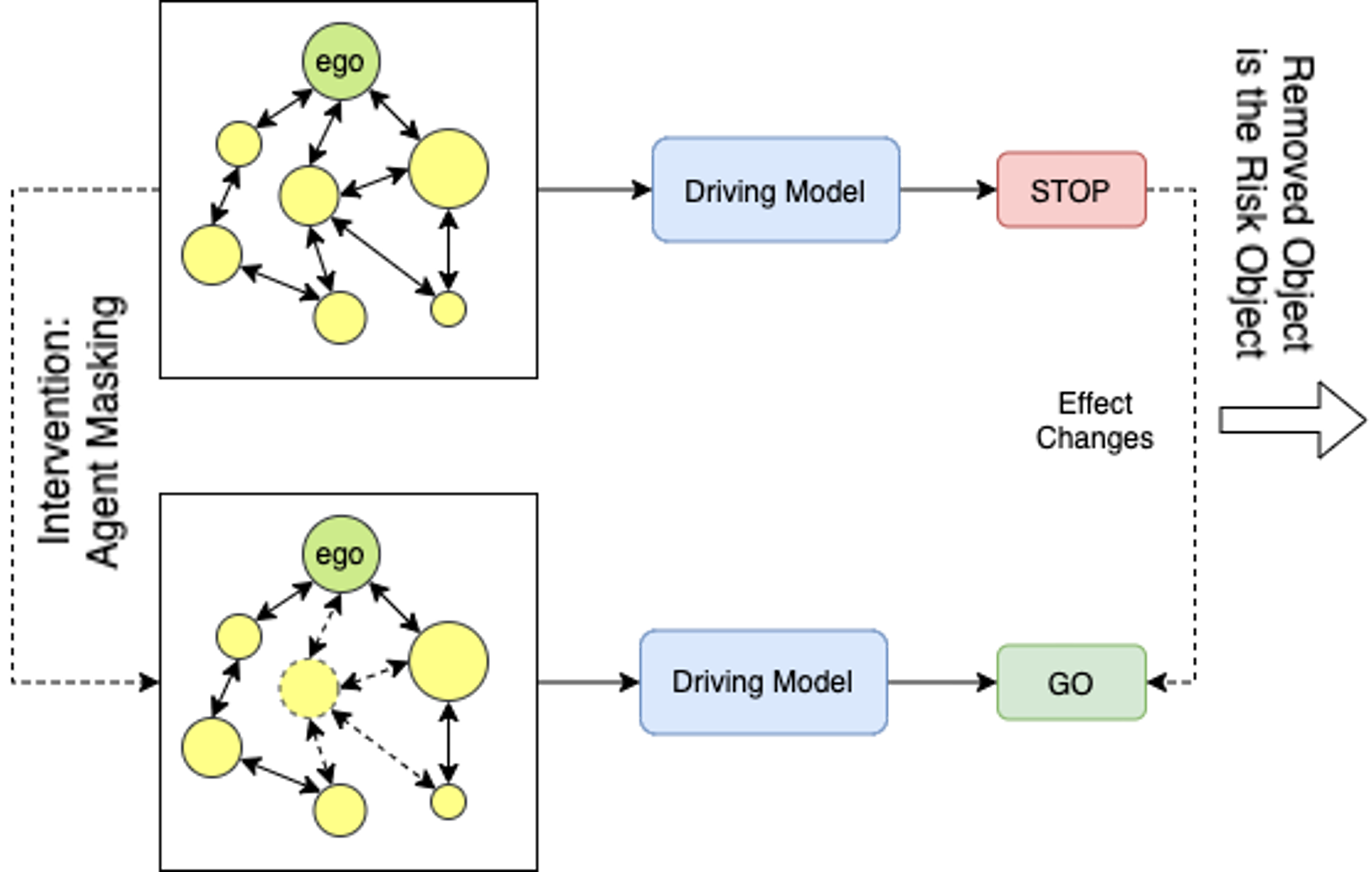}}%
  \caption{Risk-identification using a two-step inference framework. In this scenario, intervening the input by removing the agent changes the driver's behavior (effect) from \textit{Stop} to \textit{Go}, indicating the omitted object is the risk object or the cause for \textit{Stop}. In our framework, this intervention can be achieved by omitting the desired node from the traffic interaction graph.}.
  \label{fig:framework}
\end{figure}

In this section, we discuss the task of risk object identification. Our proposed framework is designed to learn the spatio-temporal interactions in traffic, and we explain how it is an appropriate fit for this task.

\subsection{Problem Statement}
We use the driver-centric definition of risk to tackle the task of risk object identification or detecting the traffic agent that has the most significant influence over the ego's tactical behavior for a given scenario \cite{li2020make}. Due to the limited annotations of real-world risk object bounding boxes, we choose to experiment with the cause-effect formulation of this task.

\subsection{Causal Inference Setup}
Given a traffic scenario $\Img^{[1..N]}_{1:\T}$ of $\T$ frames with $N$ unique road agents, the proposed framework $f$ is used to perform a two-stage inference: 
\begin{itemize}
    \item predict the high-level driving decision as \textit{Stop} or \textit{Go}
        \begin{align}
            f: \Img^{[1..N]}_{1:\T} \rightarrow \{\text{\textit{Stop}}, \text{\textit{Go}}\}
        \end{align}
    \item if the \textit{Stop} prediction confidence is high, 
        \begin{align}
            Pr\left[f\left(\Img^{[1..N]}_{1:\T}\right) = \text{\textit{Stop}}\right] \geq \delta
        \end{align}
    where $\delta$ is the thresholding hyperparameter, then the tracklet of risk object $\ro_o$ can be estimated by
        \begin{align}
            o = \argmax_{i \in [1..N]} Pr\left[f\left(\Img^{[1..N] \setminus i}_{1:\T}\right) = \text{\textit{Go}}\right] \label{eqn:argmax}
        \end{align}
        where $\Img^{[1..N] \setminus i}_{1:\T}$ refers to the manipulated input by excluding tracklet $\ro_i$ from the original input $\Img^{1..N}_{1:\T}$, and $Pr\left[f\left(\Img^{[1..N] \setminus i}_{1:\T}\right) = \text{\textit{Go}}\right]$ refers to the risk score $s_i$ of agent with tracklet $\ro_i$. 
\end{itemize}
We note that this is a simplistic formulation to detect risk objects in critical \textit{Stop} scenarios where some road agent might have a particularly high influence causing the ego's behavior to be \textit{Stop}. The probability maximizing tracklet $\ro_o$ can be obtained from Equation \eqref{eqn:argmax} by performing iterative manipulations for each of the $N$ road agents, one at a time. The risk evaluation is weakly supervised and only requires frame-level \textit{Stop/Go} annotations for training.

\subsection{Our Approach}
The design of our proposed model allows for trivial manipulation of the traffic scenario with respect to each road agent by masking out its attributes before the graph modeling step. The spatio-temporal graph thus generated should closely represent an actual scenario without the particular agent. This avoids the artifacts that are introduced from masking directly in the input video frames \cite{li2020make}. We show improved results compared to the baseline in the next section.

We realize a drawback in the above causal inference setup towards safe risk assessment. Let's consider a scenario with multiple pedestrians crossing in front of the ego, and the original tactical behavior is correctly classified as \textit{Stop}. 
In such a case, unless the crossing pedestrians are removed all at once, the driving model's prediction shouldn't change from \textit{Stop} to \textit{Go}, i.e., any pedestrian's individual risk score $s_i$ needn't be very high. Hence, we can cherry-pick such scenarios from the dataset and identify the corresponding \textit{risk groups} based on their high interaction values with respect to the ego node in the spatio-temporal graph $\G_{1:\T}$. For some of these scenarios, we perform the causal inference step with a single manipulation for the entire risk group and show the risk scores in this visualization.

\section{Experiments}

\begin{table}
\begin{center}
\begin{tabularx}{\columnwidth}{|X|X|X|X|X|}
\hline
\multirow[b]{2}{*}{Method} & \multicolumn{4}{|c|}{$m A c c$ (\%)} \\
\hline
 & $\begin{array}{c}\text { Crossing } \\ \text { Vehicle }\end{array}$ & $\begin{array}{c}\text { Crossing } \\ \text { Pedestrian }\end{array}$ & $\begin{array}{l}\text { Parked } \\ \text { Vehicle }\end{array}$ & Congestion \\
\hline
Random Selection & $15.1$ & $7.1$ & $6.4$ & $5.5$ \\
\hline
Xia et al. \cite{Xia2019} & $16.8$ & $8.9$ & $10.0$ & $21.3$ \\
\hline
Wang et al. \cite{wang2019deep} & $36.5$ & $21.2$ & $20.1$ & $8.9$ \\
\hline
Kim et al. \cite{kim2017interpretable} & $41.9$ & $21.5$ & $34.6$ & $62.7$ \\
\hline
Li et al. \cite{li2020make} & \cellcolor{red!25} $43.0$ & \cellcolor{orange!25}$27.0$ & \cellcolor{orange!25}$39.8$ & \cellcolor{orange!25}$81.0$ \\
\hline
\textbf{Ours} & \cellcolor{orange!25} ${42.7}$ & $\cellcolor{red!25}{28.6}$ & $\cellcolor{red!25}{41.3}$ & $\cellcolor{red!25}{\mathbf{87.9}}$ \\
\hline
\end{tabularx}
\end{center}
\caption{Comparison with baseline methods. The methods with $*$ are reimplemented by Li et al. \cite{li2020make} . The best and second performances are colored in \textcolor{red!75}{red} and \textcolor{orange!75}{orange}, respectively.}
\label{tab:res}
\end{table}

\subsection{Dataset}
For the task of tactical behavior prediction, we evaluate our approach on the HDD dataset \cite{ramanishka2018toward}. This dataset is particularly challenging and facilitates research in learning driver behavior in real-life settings. It consists of 104 hours of actual human driving data recorded in the San Francisco Bay Area, captured using a vehicle equipped with various sensors. The videos from the dataset offer driver-centric perspectives, and they are augmented with GPS coordinates and CAN bus sensor data overlaid on the front-facing camera stream.

The HDD dataset provides driver-centric videos with frame-level annotations of tactical driver behavior and their corresponding causes. The dataset covers diverse road scenarios, including urban, suburban, and highway environments.

Each frame in the dataset is labeled with a 4-layer representation to describe the tactical driver behaviors. Among these layers, the \textbf{Goal-oriented action} layer (e.g., left turn and right lane change) and the \textbf{Cause} layer (e.g., stop for crossing vehicle) consist of actions with interactions. We utilize the labels from these two layers to analyze the effectiveness of our proposed interaction modeling framework.

For the task of causal risk object identification, we also use the HDD dataset, leveraging the layers of \textbf{Stimulus-driven action} with labels such as \textit{Stop} and \textit{Deviate}, and the \textbf{Cause} layer denoting the reason for the associated stimulus-driven behavior. The dataset includes scenarios labeled with six types of \textbf{Cause}, namely \textit{Stopping for Congestion, Stopping for Crossing Vehicle, Deviating for Parked Vehicle, Stopping for Pedestrian, Stopping for Sign}, and \textit{Stopping for Red Light}. For evaluating our proposed risk object identification framework, we select the first four scenarios, as our approach is based on interactions between dynamic road agents only. To train our driving model for risk assessment, we use the frame-level driver behavior labels (\textit{Go} and \textit{Stop}). For evaluating the task of risk identification, we utilize the object-level annotations \cite{li2020make} provided in the dataset.

\subsection{Evaluation metrics}
We evaluate the performance of our model using mean accuracy ($mAcc$). Accuracy is computed as the number of correct predictions over the number of ground truth samples. A prediction is considered accurate if the Intersection over Union (IoU) score is greater than a certain threshold. We compute $mAcc$ as the average
accuracy at 10 IoU thresholds, uniformly distributed in the range of $0.5 - 0.95$.

\subsection{Implementation Details}
For tactical behavior prediction, we use 20-frame clips with resolution $200\times356$ at $3$ fps as the input to the framework. We use the I3D network \cite{Carreira_2017_CVPR} pre-trained on the Kinetics action recognition dataset \cite{kay2017kinetics} as the backbone for encoding the global video features. The intermediate video feature map from the \texttt{Mixed\_3c} layer of I3D is used in the ROIAlign layer to obtain the appearance features. In our implementation, we use the appearance embedding dimension $D=256$. For encoding the positional relations using $\gamma(.)$, we choose $d=5$, $k=30$ and
follow \cite{tancik2020fourier} to choose $\sigma=10$. We stack three layers of ST-GCN, i.e., the pair of equations \eqref{eqn:scn} and \eqref{eqn:tcn} with temporal span $\tau=3$. The last layer of ST-GCN is followed by the classifier that includes average pooling and fully connected layers to obtain the classification logits, which are further converted into class probabilities using the softmax layer. Each layer of a neural network is followed by a layer of batch normalization. We use Adam as the optimizer with default parameters and an initial learning rate of $0.001$. We are implementing the framework on PyTorch \cite{paszke2019pytorch} and performing all experiments on a node with 4 Nvidia Tesla M40 GPUs. 

\begin{figure}[!t]
  \centering
  \begin{subfigure}{0.45\linewidth}
      {\includegraphics[width=\linewidth]{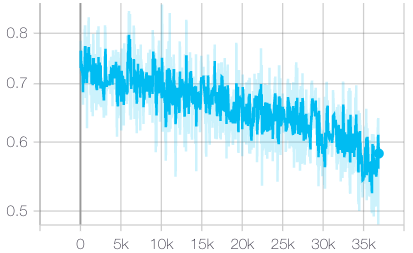}}
      \caption{training loss curve}
  \end{subfigure}
  \hfill
  \begin{subfigure}{0.45\linewidth}
      {\includegraphics[width=\linewidth]{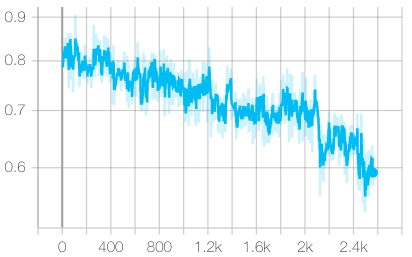}}
      \caption{validation loss curve}
  \end{subfigure}
  \caption{We present the training and validation loss curves obtained from the training for 40 epochs.}
  \label{fig:framework}
\end{figure}

\section{Quantitative Evaluation}

Risk object identification is a relatively new area of research in the autonomous driving community. We compare with previous approaches in this area to select important/risk objects. The comparison with our method is shown in TABLE \ref{tab:res}. We represent the performance of our approach on top of the results of several baseline methods reported in \cite{li2020make}. 

\noindent \textbf{Random Selection:} This baseline method identifies a risk object by randomly selecting an object detected in a frame without considering any surrounding information. The Driver's Attention Prediction method uses a pre-trained model \cite{Xia2019} to predict the driver's gaze attention maps for each frame. The method calculates the average attention weight of each detected object region and designates the object with the highest attention weight as the risk object, indicating where the driver's gaze is focused. The model is trained from a pre-trained model of the BDD-A dataset \cite{Xia2019} since the HDD dataset lacks human gaze information. Compared to Random Selection, this method delivers slightly superior performance. 

\noindent \textbf{Pixel-level Attention + Causality Test:} Kim et al. \cite{kim2017interpretable} devised a causality test for identifying regions that influence a network's output. To achieve this, they created a pixel-level attention map using an end-to-end driving model to extract particles based on attention value over an input image. These particles were then clustered to create convex hull region proposals. An RGB image was evaluated repeatedly by the trained model to conduct a causality test, and the region that resulted in the largest decrease in prediction performance was deemed the risk object.

\noindent \textbf{Object-level Attention Selector}: Wang et al. \cite{wang2019deep} developed a driving model that concentrates on objects by learning attention weights at the object level, enabling the identification of risky objects. We were influenced by their methodology and modified the message passing in our driving model to integrate object-level attention, following which we retrained the model. Subsequently, we evaluated the model's accuracy by selecting the object with the highest attention weight in four distinct scenarios.

\noindent \textbf{Who Make Drivers Stop?:} Li et al. \cite{li2020make} suggests a new definition of risk that focuses on objects that impact a driver's behavior. Motivated by design by Wang et al. \cite{wang2019deep}, the authors modify the message passing in the driving model to be object-level attention and retrain their model. They propose a new task called risk object identification, which aims to solve a cause-effect problem. The authors introduce a two-stage risk object identification framework that relies on causal inference and the object-level manipulable driving model.

We compare the performance of these models with our approach. As indicated in Table \ref{tab:res}, Random Selection and Driver's Attention Prediction models are inferior in comparison to other models. The Object-level Attention Selector approach achieves better results but performs poorly on congestion tasks. Pixel-level Attention and Causality Test achieve better performance, but still worse than Who Make Drivers Stop in all the tasks. Our model achieves top results in 3 out of 4 tasks. For the crossing vehicle setting, our result is very close to that of the top model.

\section{Conclusion, Limitations, and Future Work}

We've introduced a new autonomous driving framework that uses spatio-temporal traffic graphs to mirror human driving behavior, leading to an increased ability to identify risky interactions. Our results outperform all existing baseline models.

Our future work will involve integrating psychology-based traffic interaction algorithms and examining multiple data sources, such as vehicle, driver biometric, and environment sensors, to boost risk prediction accuracy. We also plan to develop complex models involving deep learning and reinforcement learning and test these models in real-world scenarios to validate their efficacy and discover any limitations. In addition, we aim to cultivate explainable AI techniques to enhance the interpretability of driver risk prediction models, promoting broader adoption and trust.

Our model offers the potential to increase autonomous vehicle safety by helping them predict human-like driving behaviors, reducing accident risks. It can be used in driver assistance systems for real-time risk assessment and in formulating insurance policies by assessing driving behavior risks. The promise held by our framework extends to advancements in autonomous driving and safer, more efficient transportation systems.

{\small
\bibliographystyle{ieee_fullname}
\bibliography{references}
}

\end{document}